\documentclass[10pt,twocolumn,letterpaper]{article}

\usepackage{authblk}

\usepackage{cvpr}              

\usepackage{graphicx}
\usepackage{amsmath}
\usepackage{amssymb}
\usepackage{booktabs}
\usepackage[dvipsnames]{xcolor}
\usepackage{slashbox}

\usepackage{tabularx,booktabs}

\newcolumntype{C}{>{\centering\arraybackslash}X} 
\usepackage[pagebackref,breaklinks,colorlinks]{hyperref}

\makeatletter
\renewcommand\AB@affilsepx{\quad \protect\Affilfont}
\makeatother

\usepackage[capitalize]{cleveref}
\crefname{section}{Sec.}{Secs.}
\Crefname{section}{Section}{Sections}
\Crefname{table}{Table}{Tables}
\crefname{table}{Tab.}{Tabs.}

\def\cvprPaperID{2} 
\def\confName{CVPR}
\def\confYear{2023}

\title{U2RLE: Uncertainty-Guided 2-Stage Room Layout Estimation}

\begin{document}

\author[1]{Pooya Fayyazsanavi$^{*\dagger}$}
\author[2]{Zhiqiang Wan$^*$}
\author[2]{Will Hutchcroft}
\author[2]{Ivaylo Boyadzhiev}
\author[2]{Yuguang Li}
\author[1]{Jana Kosecka}
\author[2]{Sing Bing Kang}

\affil[1]{George Mason University}
\affil[2]{Zillow Group}

\maketitle

\def\thefootnote{*}\footnotetext{Equal contribution.}\def\thefootnote{\arabic{footnote}}
\def\thefootnote{$\dagger$}\footnotetext{This work was done when Pooya Fayyazsanavi was an intern at Zillow.}\def\thefootnote{\arabic{footnote}}

\def\thefootnote{$1$}\footnotetext{\{pfayyazs,kosecka\}@gmu.edu}\def\thefootnote{\arabic{footnote}}
\def\thefootnote{$2$}\footnotetext{\{zhiqiangw,willhu,ivaylob,yuguangl,singbingk\}@zillowgroup.com}\def\thefootnote{\arabic{footnote}}

\begin{abstract}
While the existing deep learning-based room layout estimation techniques demonstrate good overall accuracy~\cite{Sun_2019_CVPR}, they are less effective for distant floor-wall boundary. To tackle this problem, we propose a novel uncertainty-guided approach for layout boundary estimation introducing new two-stage CNN architecture termed U2RLE. The initial stage predicts both floor-wall boundary and its uncertainty and is followed by the refinement of boundaries with high positional uncertainty using a different, distance-aware loss. Finally, outputs from the two stages are merged to produce the room layout. 
Experiments using ZInD~\cite{ZInD} and Structure3D~\cite{Zheng2020Structured3DAL} datasets show that U2RLE improves over current state-of-the-art, being able to handle both near and far walls better. In particular, U2RLE outperforms  current state-of-the-art techniques for the most distant walls.

\end{abstract}

\section{Introduction}

A lot of progress has been made in image-based room layout estimation (RLE). While deep learning techniques applied to RLE have been very effective, they are evaluated using datasets of rooms that are mostly small and cuboid-shaped~\cite{zhang2014panocontext, armeni2017joint}. However, ZInD~\cite{ZInD}, which is a large-scale dataset of real homes, shows that real rooms are large and have more complex layouts.




\begin{figure}[!t]
    \centering
    \subfloat[Room only with close walls.]{\includegraphics[width=2.0in, height=1.2in]{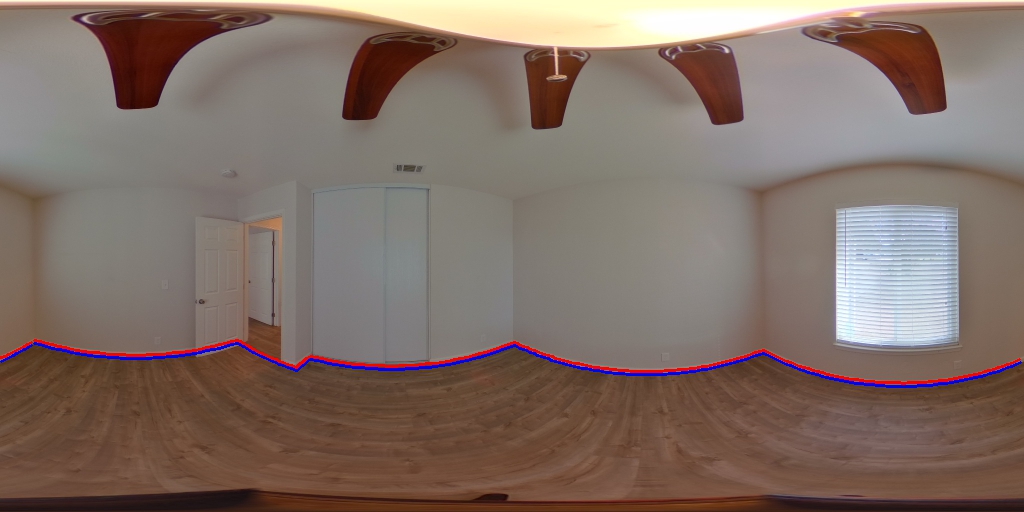}
    \label{fig:pano_view_close_wall}
    }
    \subfloat[\textcolor{red}{GT}: $ 12.4 m^2 $; \\ \textcolor{blue}{GT+3 pixel error}: $ 11.4 m^2 $]{\includegraphics[width=1.4in, height=1.3in]{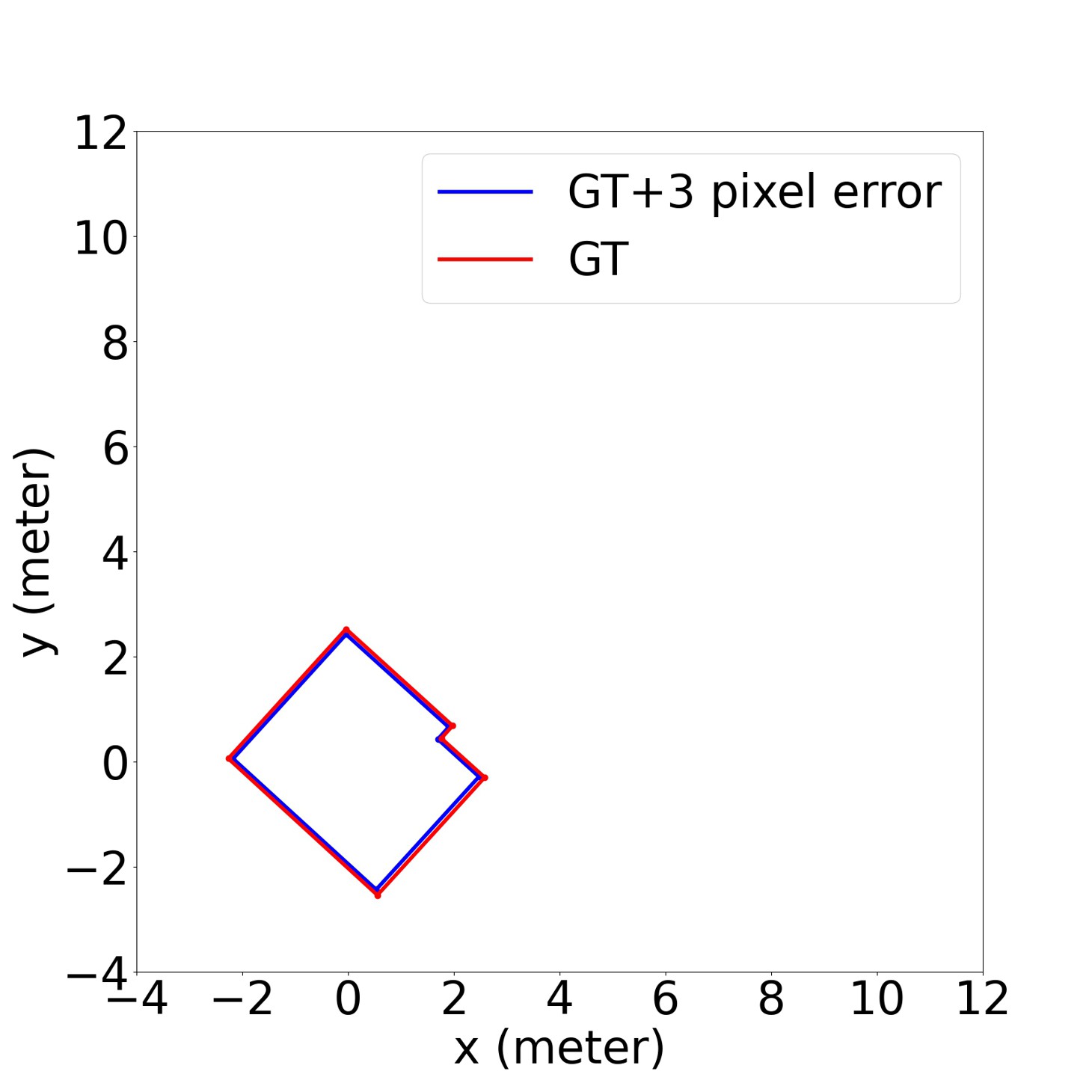}
    \label{fig:top_down_view_close_wall}
    }
    
    \subfloat[Room with distant walls.]{\includegraphics[width=2.0in, height=1.2in]{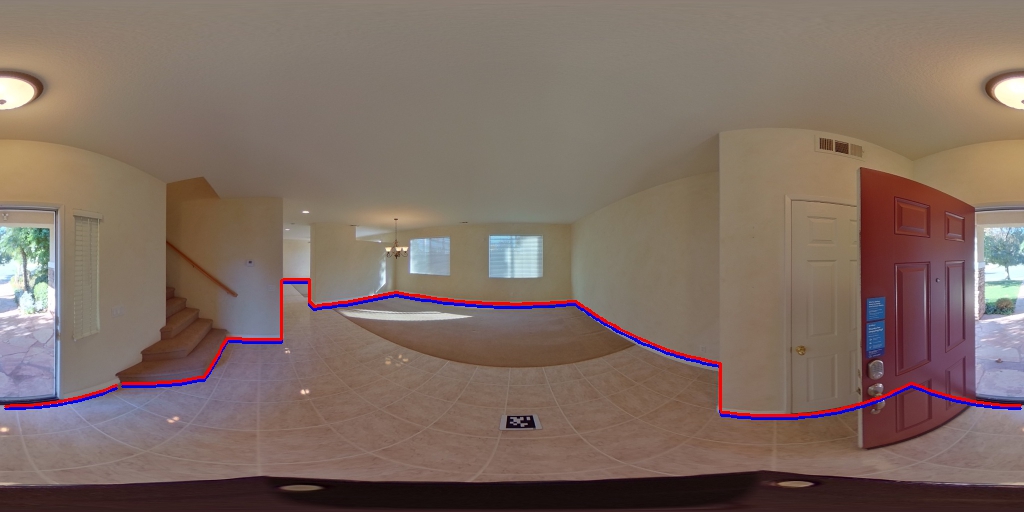}
    \label{fig:pano_view_distant_wall}
    }
    \subfloat[\textcolor{red}{GT}: $ 45.1 m^2 $; \\ \textcolor{blue}{GT+3 pixel error}: $ 38.6 m^2 $]{\includegraphics[width=1.4in, height=1.3in]{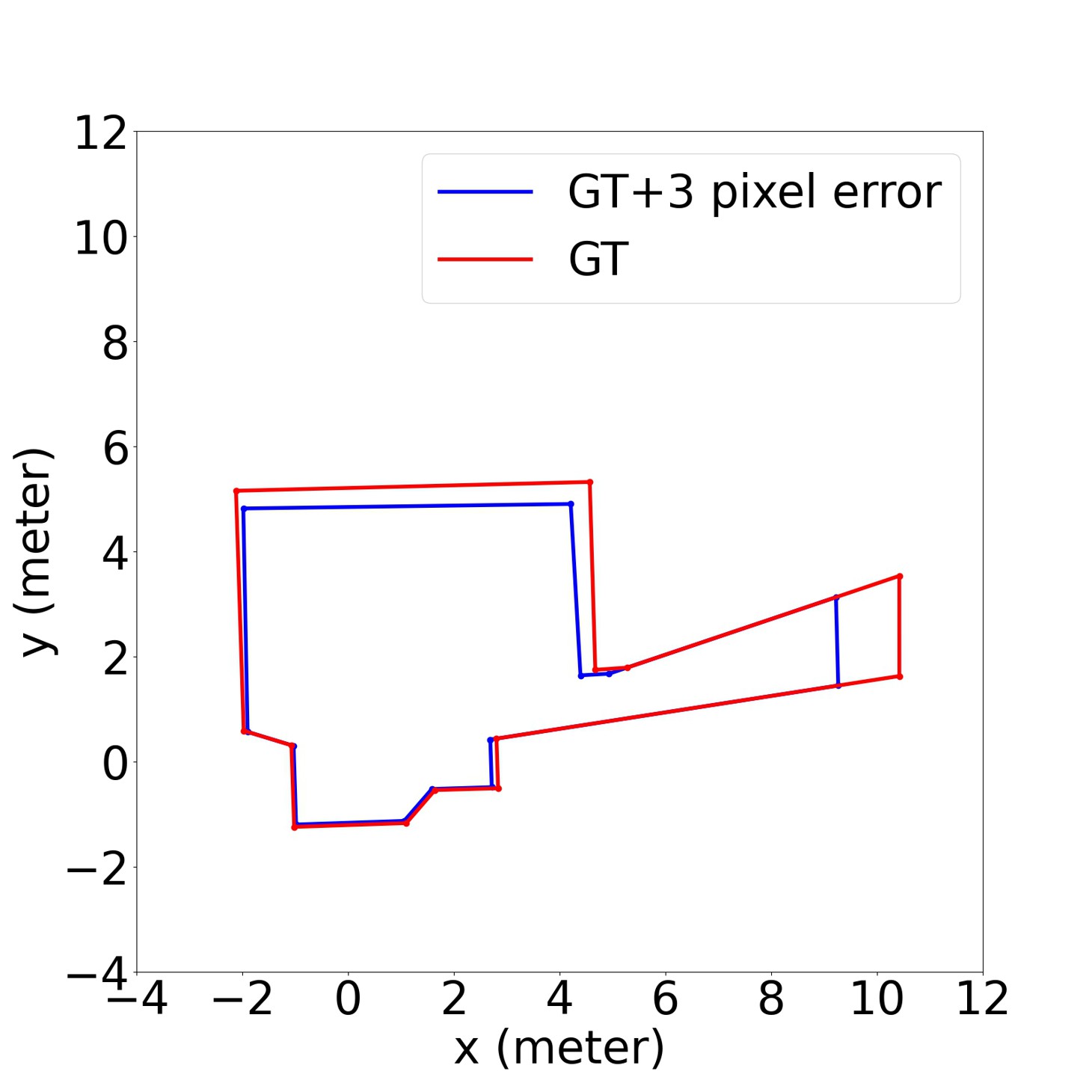}
    \label{fig:top_down_view_distant_wall}
    }
    \caption{The same pixel error from a floor point maps to a larger depth error when the point is further away from the camera. \textcolor{red}{GT} layout is in \textcolor{red}{Red} while the simulated \textcolor{blue}{perturbed} layout is in \textcolor{blue}{Blue}. (b) and (d) share aligned coordinate system. }
    \label{fig:motivation}
\end{figure} 

The challenge of distant wall depth estimation can be seen clearly from the relationship between pixel and depth errors in Fig.~\ref{fig:motivation}. We simulate a \textcolor{blue}{perturbed layout} by shifting the vertical position of each room corner of the \textcolor{red}{GT layout} by 3 pixels. With an image height of 512 pixels, 3 pixels corresponds to 0.59\% relative error in image-space, for both near and distant walls, alike; however, this results in significantly larger depth error for distant walls. For the room with only close walls (Fig.~\ref{fig:top_down_view_close_wall}), the area of the \textcolor{red}{GT layout} and the \textcolor{blue}{perturbed layout} are $ 12.4m^2 $ and $ 11.4m^2 $, respectively, resulting in just $1m^2$ error. For a room with distant walls, the area of the \textcolor{red}{GT layout} and the \textcolor{blue}{perturbed layout} are $ 45.1m^2 $ and $ 38.6m^2 $, respectively, an error of $ 6.5m^2 $. As we can see, distant walls have a big impact on estimating the area of a room. For real-world applications such as real estate, where the floorplan size is a major factor in pricing, this magnitude of the absolute error is undesirable, pointing to the need for new methods to address this challenge. 

While recent large-scale datasets~\cite{ZInD, Zheng2020Structured3DAL} provide large annotated spaces with more distant walls, dataset imbalance challenges accurate estimation. For example, in ZInD~\cite{ZInD}, nearly 90\% of the walls are within 4 meters of the camera while the remaining walls can extend out to 10+ meters. Many current models are trained to minimize the loss averaged across all image columns; imbalance may lead to a model that does not prioritize the feature resolution and granularity necessary to accurately estimate distant walls.

\begin{figure*}[t]
\begin{center}
   \includegraphics[width=0.90\linewidth]{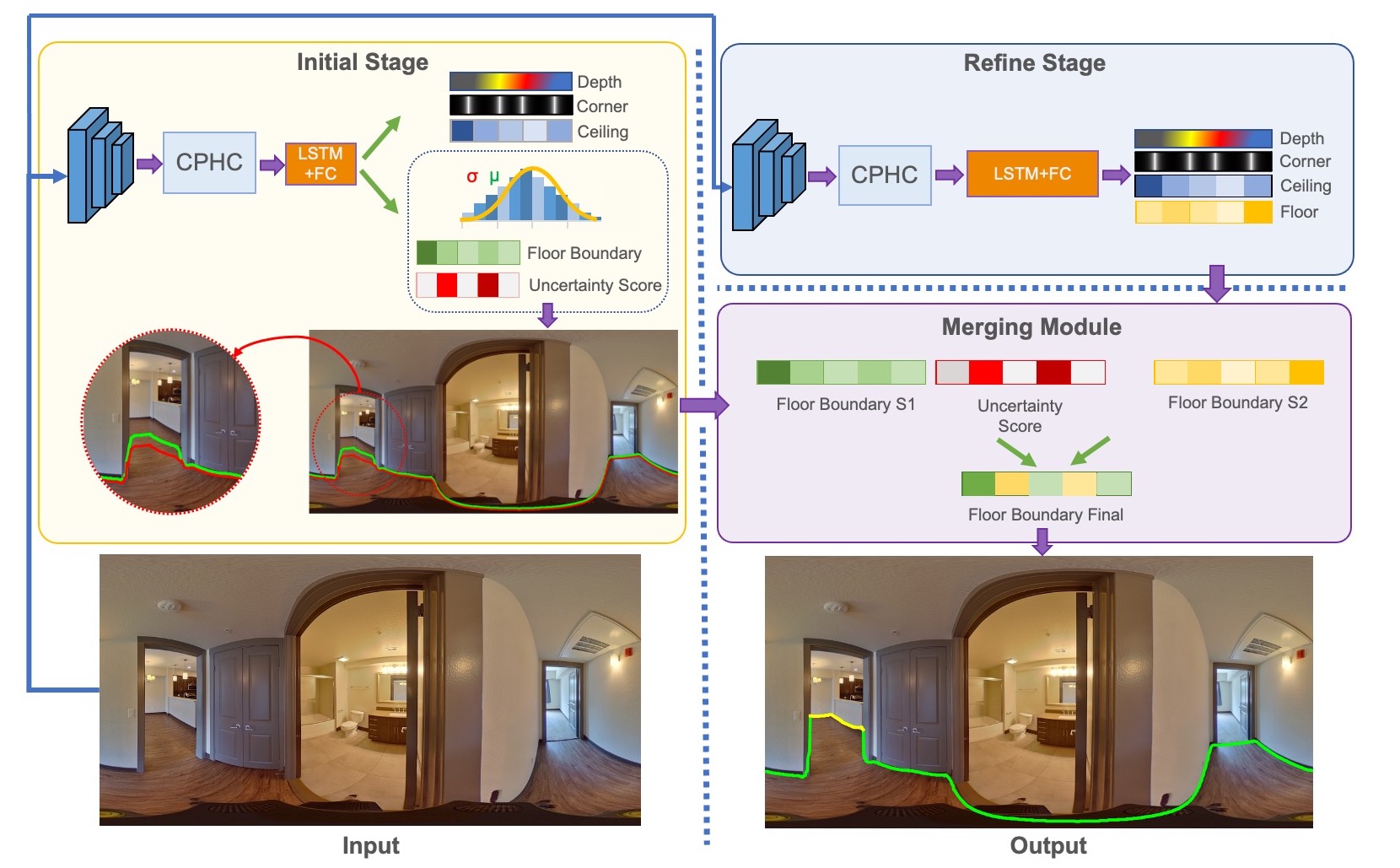}
\end{center}
    \caption{Overview of U2RLE. It has three main components: initial stage, refinement stage, and merging module. In the initial stage, the model predicts the boundary and estimates its positional uncertainty. In the initial stage, the gap between the \textcolor{green}{green} boundary and \textcolor{red}{red} boundary represents the uncertainty score. The refinement stage focuses on distant regions using distance-aware loss. Finally, the merging module combines the predictions from both stages using the uncertainty score. In the merging module, the uncertain parts indicating by \textcolor{red}{red}  will be replaced by the floor-wall boundary prediction from the refinement stage. }
    \label{arch}
\end{figure*}

To address these challenges, we propose a novel two-stage architecture. The first (initial) stage is designed to predict the wall position, alongside a measure of uncertainty. The second (refine) stage focuses on distant regions using distance-aware loss. Finally, guided by the predicted uncertainty score, the outputs of these two stages are merged to generate the room layout. This two-stage architecture allows us to design specific loss functions and data augmentation for close vs. distant walls; it handles the imbalance in the wall depth distribution. 



Our contributions are:
\begin{itemize}
    \item Uncertainty-guided two-stage (initial and refine) room layout estimation model, achieving SOTA performance on multiple datasets.
    \item Novel loss function to predict floor-wall boundary uncertainty in the initial stage. 
    \item New Channel-Preserving Height Compression (CPHC) module that compresses features along its height.
    \item New distance-aware loss to re-weight the influence of distant walls in the refinement stage. 
\end{itemize}

\section{Related Work}


\paragraph{Single-view Room Layout Estimation.}
Single-view room layout estimation has been an active area of research for the past decade~\cite{Mathew2020ReviewOR}. A good amount of work has been done to generate partial room layouts from a single perspective image. Learning-based approaches \cite{ren2016coarse, lee2017roomnet, yan20203d} have been proven more effective compared to geometry-based ones \cite{flint2010dynamic, hedau2009recovering}. PanoContext~\cite{zhang2014panocontext} is the first method to demonstrate the effectiveness of room layout estimation using a single 360$^\circ$ equirectangular panorama image. It generates the room layout by estimating per-pixel normals of overlapping perspective images converted from a single panorama. LayoutNet~\cite{zou2018layoutnet} demonstrated the benefits of using encoder-decoder CNN-based architecture to directly operate on the panoramic image; layouts are predicted by estimating the probability maps of room boundary map and corners, followed by a 3D layout optimization.

HorizonNet~\cite{Sun_2019_CVPR} simplified layout estimation tasks by generating a 1D representation of the floor-wall, ceiling-wall, and wall-wall boundary positions per image column; it uses a bidirectional LSTM (Bi-LSTM) module to regress across the panorama. Since then, many approaches are based on the HorizonNet architecture. HoHoNet\cite{SunSC21} re-designed the feature extractor with the Efficient Height Compression (EHC) module followed by multi-head self-attention (MSA) to generate Latent Horizontal Feature (LHFeat). They demonstrated good performance using LHFeat to model per-column modality for room layout reconstruction as well as dense predictions such as depth estimation and semantic segmentation. LED2Net\cite{Wang_2021_CVPR} adopted HorizonNet’s backbone and applied differentiable depth rendering to incorporate the room polygon geometry information into their end-to-end model. LGT-Net\cite{jiang2022lgt} used a novel transformer architecture as their sequence processor to improve the spatial identification ability for the panorama. LED2Net and LGT-Net both achieve state-of-the-art performance overall; the recent focus of layout estimation study has placed more emphasis on solving complex, non-Manhattan room shapes, which are found more frequently in real-life, large-scale datasets such as ZinD\cite{ZInD} and CG datasets such as Structure3D\cite{Zheng2020Structured3DAL}.

While using equirectangular panorama images as inputs, DuLaNet\cite{Yang:2019:DuLa-Net}, AtlantaNet\cite{Pintore2020AtlantaNetIT}, and PSMNet\cite{wang2022psmnet} also leverage the Equirectangular-to-Perspective (E2P) transform to generate a binary segmentation map as the floor plan. AtlantaNet demonstrated an advantage of modeling non-Manhattan and complex room layouts. DMHT\cite{zhao20223d} alternatively applied a learnable Hough Transformation Block on cubemap tiles from panoramic images and claimed SOTA performance. 




\paragraph{Multi-view Room Layout Estimation.}
Additional view-points can be used to improve layout estimation for complex, large rooms~\cite{wang2022psmnet, gprnet_2022}. Furthermore, recent work~\cite{solarte2022mlc} has used multi-view consistency, in a self-supervised approach, to improve single-room layout estimation.


\paragraph{Uncertainty Estimation.} 
Several approaches for depth estimation in the past used iterative techniques to improve 
the depth estimates either using additional cues, such as plane normals~\cite{Bae_IronDepth_arxiv2022}. General framework for uncertainty estimation in Deep Learning has also been adopted in the past for geometric regression problems~\cite{Loquercio_Uncertainty_RAL2020}. These class of methods uses Monte-Carlo (MC) sampling to estimate uncertainty, where the samples are generally computed using an ensemble of neural networks. The application of uncertainty to the problem of estimating room layouts is however new. Here we use simpler yet effective approach where both the mean and the variance are directly predicted from data.

\section{System Overview}

We present U2RLE, a novel two-stage approach to estimate the layout of a room using a single 360$^\circ$ equirectangular panorama. Figure \ref{arch} shows the overall architecture of our method. We describe the initial stage in section \ref{firststage}. The goal of the initial stage is to provide initial boundary and uncertainty predictions. The goal of the refinement stage, which we describe in section \ref{secondstage}, is to improve the far away region using a novel distance-aware loss function. Finally, in section \ref{merging}, we discuss how to utilize the predicted uncertainty score to combine predictions from both stages to get the best of both worlds, i.e. close and distant regions.



\section{Initial Boundary and Uncertainty Prediction}
\label{firststage}


The initial stage follows the approach of HorizonNet \cite{Sun_2019_CVPR}. The input image of size 512 $ \times $ 1024 $ \times $ 3 is passed through ResNet-50 to extract features. Then, we propose a new Channel-Preserving Height Compression Module (CPHC) to compress the information along height dimension, so as to obtain per column features that can be used for boundary prediction. The compressed features in the column order are then passed through bi-directional LSTM and fully-connected layers to predict the boundary and the uncertainty score for each column of the panorama, along with corners and depth. 


\subsection{Channel-preserving Height Compression}
\label{hcm}
HorizonNet\cite{Sun_2019_CVPR} proposed the height compression module (HCM) to effectively squeeze features with 2D spatial support from the backbone along the height dimension to produce a 1D horizontal feature. Many recent room layout approaches~\cite{jiang2022lgt,Wang_2021_CVPR,Chen2021WideBaselineRC} adopt HCM yielding good results. The features from block 1 $ \sim $ 4 of ResNet-50 are compressed to dimensions (channel $ \times $ height $ \times $ width) $ 32 \times 8 \times 256, 64 \times 4 \times 256, 128 \times 2 \times 256, 256 \times 1 \times 256 $. As we can see, the height dimension of block 1 $ \sim $ 3 has not been compressed to 1. In HCM, the first two dimensions (channel and height) will be reshaped into one dimension. We argue this is not the most effective way to compress the height information. Since we predict one value for each image column, it is better not to mix the height and channel dimensions. To address this issue, we propose a new Channel-Preserving Height Compression (CPHC) Module that compresses the height dimension into 1 and avoids mixing the information from feature channels and height dimension. The architecture of the proposed CPHC is shown in supplementary material. The ablation study in Tab.~\ref{tab:new_hcm} demonstrates CPHC module’s ability to enhance the performance of both initial and refinement stage.

\subsection{Uncertainty Prediction}
\label{nll}
Unlike the existing models, the initial stage not only predicts the floor-wall boundary but also the uncertainty of the prediction. Specifically, the model predicts two quantities per each column namely floor-wall boundary mean $\mu$, and floor-wall boundary standard deviation $\sigma$. 



The model is optimized by minimizing the negative log likelihood loss function for each image column $i$ with
$\mu_{i}$ and $\sigma_{i}$:\\
\begin{equation}
\label{eq:nll_loss}
\mathcal{L}_{floor}^{\mathrm{NL}}=-\sum_{i=1}^W \log \frac{1}{\sigma_i \sqrt{2 \pi}} \exp{\left(-\frac{\left(y_i-\mu_i\right)^2}{2 (\sigma_i)^2}\right)}
\end{equation}

where $y_{i}$ is the ground truth floor-wall boundary, and $W$ is the image width. During the inference, the column-wise predicted mean $\mu_i$ vector will be used as the predicted floor-wall boundary. Since the predicted uncertainty vector $\sigma_i$ is in the image-space, we need to project it to the depth-space. We begin by calculating the {\em upper bound} $\widehat{y}$ of the floor-wall boundary by adding the predicted mean vector $\mu$ to predicted standard deviation vector $\sigma$:
\begin{equation}
\widehat{y}=\mu+\sigma
\end{equation}
The predicted floor-wall boundary ($ \mu $) and the upper bound $\widehat{y}$ are projected into the depth-space omitting the index $i$ for clarity. We define the uncertainty score as the distance between the projected floor-wall boundary and the upper bound.
\begin{equation}
U_{floor}= \lvert Proj(\widehat{y}) - Proj(\mu) \rvert
\end{equation}
where $Proj(.)$ is the projection function to depth. Details about this projection function can be found in \cite{jiang2022lgt}.



\subsection{Loss Function}

The total loss is calculated as 
\begin{equation}
\label{firsttotalloss}
\mathcal{L}=\mathcal{L}_{floor}^{\mathrm{NL}}+ \mathcal{L}_{\text {ceil }} + \mathcal{L}_{\text {depth }} + \lambda \mathcal{L}_{\text {corner }} 
\end{equation}
where the negative likelihood loss $\mathcal{L}_{floor}^{\mathrm{NL}}$ of floor-wall boundary is described in Eq. \ref{eq:nll_loss}; $\mathcal{L}_{ceil}$ is $\ell_1$ loss of ceiling-wall boundary; $\mathcal{L}_{depth}$ represents the $\ell_1$ loss of wall depth; $\mathcal{L}_{corner}$ is the corner loss.

Unlike HorizonNet where Binary Cross-Entropy Loss is used as corner loss, we propose to replace it with a $\ell_1$ loss. This is because the ground truth is a continuous value instead of a binary value.

\begin{table*}
\footnotesize
\begin{tabularx}{\textwidth}{@{} l *{15}{C} c @{}}
\toprule
\backslashbox{Model}{GT Depth (m)} 
& 1 & 2 & 3 & 4 & 5 & 6 & 7 & 8 & 9 & 10 & 2$\mathrm{DIoU}$ \\ 
\midrule
HorizonNet\cite{Sun_2019_CVPR}  & \footnotesize{0.035} & \footnotesize{0.048}  & \footnotesize{0.073} & \footnotesize{0.136} &	\footnotesize{0.256} &	\footnotesize{0.367} &	\footnotesize{0.540}  &	\footnotesize{0.757} &	\footnotesize{0.964} &	\footnotesize{1.267}  &	 \footnotesize{89.80\%} \\ 


LED2-Net \cite{Wang_2021_CVPR} & \footnotesize{0.036} & \footnotesize{0.049}  & \footnotesize{0.070} & \footnotesize{0.125} &	\footnotesize{0.222} &	\footnotesize{0.343} &	\footnotesize{0.526}  &	\footnotesize{0.800} &	\footnotesize{0.903} &	\footnotesize{1.156}  &	 \footnotesize{89.95\%}  \\ 


LGT-Net\cite{jiang2022lgt}   & \footnotesize{0.035} & \footnotesize{0.051}  & \textbf{\footnotesize{0.067}} & \textbf{\footnotesize{0.123}} &	\footnotesize{0.233} &	\footnotesize{0.326} &	\footnotesize{0.501}  &	\footnotesize{0.693} &	\footnotesize{0.858} &	\footnotesize{1.145}  &	 \footnotesize{90.82\%} \\ 


Ours(U2RLE)  & \textbf{\footnotesize{0.034}} &  \textbf{\footnotesize{0.047}}  &  \footnotesize{0.072} &  \footnotesize{0.127} &	 \textbf{\footnotesize{0.229}} & \textbf{\footnotesize{0.313}} &	\textbf{\footnotesize{0.437}}  & \textbf{\footnotesize{0.614}} &	\textbf{\footnotesize{0.776}} &	\textbf{\footnotesize{0.942}}  &  \textbf{\footnotesize{91.39\%}} \\ 



\bottomrule
\end{tabularx}
\caption{Mean depth error (m) $ \downarrow $ on ZinD dataset.}
\label{tab:zind}
\end{table*}

\begin{table*}
\footnotesize
\begin{tabularx}{\textwidth}{@{} l *{10}{C} c @{}}
\toprule
\backslashbox{Model}{GT Depth (m)} 
& 1 & 2 & 3 & 4 & 5 & 6 & 7 & 8 & 9 & 10 & 2$\mathrm{DIoU}$ \\ 
\midrule
HorizonNet\cite{Sun_2019_CVPR}  & \footnotesize{0.028} & \footnotesize{0.036}  & \footnotesize{0.073} & \footnotesize{0.148} &	\footnotesize{0.245} &	\footnotesize{0.344} &	\footnotesize{0.688}  &	\footnotesize{0.891} &	\footnotesize{1.423} &	\footnotesize{1.443}  &  \footnotesize{92.63\%} \\ 


LED2-Net\cite{Wang_2021_CVPR}  & \footnotesize{0.038} & \footnotesize{0.045}  & \footnotesize{0.086} & \footnotesize{0.158} &	\footnotesize{0.233} &	\footnotesize{0.298} &	\footnotesize{0.485}  &	\textbf{\footnotesize{0.527}} &	\footnotesize{1.152} &	\footnotesize{1.053}  &  \footnotesize{92.15\%} \\

LGT-Net\cite{jiang2022lgt}   & \footnotesize{0.034} & \footnotesize{0.039}  & \footnotesize{0.074} & \footnotesize{0.142} &	\footnotesize{0.232} &	\textbf{\footnotesize{0.257}} &	\footnotesize{0.516}  &	\textbf{\footnotesize{0.558}} &	\footnotesize{1.073}  &	\footnotesize{1.155}  & \footnotesize{92.81\%} \\ 



Ours(U2RLE)  &  \textbf{\footnotesize{0.027}} &   \textbf{\footnotesize{0.031}}  &   \textbf{\footnotesize{0.059}} &   \textbf{\footnotesize{0.128}} &	   \textbf{\textbf{\footnotesize{0.217}}} &  \footnotesize{0.268} &	 \textbf{\footnotesize{0.505}}  & \footnotesize{0.575} &	\textbf{\footnotesize{0.992}} &	 \textbf{\footnotesize{0.877}} & \textbf{\footnotesize{93.73\%}} \\ 


\bottomrule
\end{tabularx}
\caption{Mean depth error (m)  $ \downarrow $  on Structure3D dataset.}
\label{tab:s3d}
\end{table*}

\section{Fine-grained Distance-focused Prediction}
\label{secondstage}
Similar to the initial stage, ResNet-50 is used to extract features and the proposed CPHC compresses the extracted features. Then,  bi-directional LSTM and fully-connected layers are used to predict the floor-wall boundary, the depth, the ceiling-wall boundary, and the corners.

\subsection{Data Augmentation}
In the refinement stage, we intend to focus on distant boundaries.
Since these regions are located far from the camera, they are usually low resolution and blurry. The top-down view further magnifies prediction errors for these regions in comparison to close walls, as shown in Figure~\ref{fig:motivation}. Furthermore, in all current datasets the distant boundaries are among the least representative samples.

To enable the model to train on more of these samples, we first perform pano-stretch data augmentation \cite{Sun_2019_CVPR} and stretch each sample's layout to a greater distance during training. As opposed to the initial stage, which pushes the boundary points to both the close and far regions, we only use the data augmentation to stretch the layout to represent the far regions. In addition, different parameters were used to magnify the stretch. As with \cite{Sun_2019_CVPR,Wang_2021_CVPR,jiang2022lgt}, we also only augment the data on the x and z axes only:
\begin{equation}
\begin{aligned}
&x^{\prime}=k_x \cdot x=k_x \cdot d \cdot \cos (v) \cdot \cos (u) \\
&z^{\prime}=k_z \cdot z=k_z \cdot d \cdot \cos (v) \cdot \sin (u)
\end{aligned}
\end{equation}
where $x,z$ is the 3D coordinate projected on top down-view from panorama image coordinate $u, v$. $ d $ is the depth. $k_x$ and $k_y$ is a random value picked from a uniform distribution of range $[1,2.5]$.

\subsection{Distance-aware Loss Function}
\label{distant_aware_loss}

To force the model to focus more on the distant walls, we propose a distance-aware loss which contains two separate loss functions. The first loss function emphasizes depth prediction on projected top-down view as follows:
\begin{equation}
\label{depth_loss}
\mathcal{L}_{\mathrm{depth}}= \sum_{i=1}^W\left|\widehat{d_i} - d_i \right|
\end{equation}
where $\widehat{d}$ is the predicted depth and $ d $ is the ground truth depth.

To focus more on the distant regions we also re-weight the floor boundary loss based on the distance as follow: 
\begin{equation}
\label{topdownloss}
\mathcal{L}_{\mathrm{floor}}= \sum_{i=1}^W \left|\widehat{y_i}-y_i\right|*d_i
\end{equation}
where $ \widehat{y} $ represents the predicted floor-wall boundary while $ y $ is the ground truth floor-wall boundary.

For the refinement stage, the total loss function is calculated as follows:

\begin{equation}
\label{secondtotalloss}
\mathcal{L}=\mathcal{L}_{\text {floor}}+ \nu \mathcal{L}_{\text {depth}} + \lambda \mathcal{L}_{\text {corner }}+\mathcal{L}_{\text {ceil }}
\end{equation}
where $\lambda, \nu \in \mathbb{R}$ are hyper-parameters designed to balance the effect of each components.
Similar to the initial stage, ceiling loss $\mathcal{L}_{ceil}$ and corner loss $ \mathcal{L}_{\text {corner }} $ use $\ell_1$ loss.

\section{Uncertainty-guided Merging}
In order to determine the final floor-wall boundary, we considered the model predictions as prior information for the merging module. The predicted uncertainty score $\sigma$ and estimated distance (from initial floor boundary) are used to determine the output of the merging module. As described in section \ref{nll}, the initial stage uncertainty scores can be projected into the top-down view. Projected uncertainty scores indicate the level of uncertainty in each column. Therefore, the refinement stage results can be merged based on the score. According to our empirical analysis, we used 0.2 as the threshold for uncertainty $U_{floor}$ across all the experiments. Also, to further refine the predictions
we only combine the results if the predicted wall is distant. Consequently, fewer outliers (close walls) will be merged from the refinement stage into the final prediction. We therefore use the estimated distance as a second parameter, which will be merged if the distance is greater than 5 meters. The module finally outputs the final prediction with size of $W$.

\label{merging}

    

    

\section{Experiments}
We report results of our approach on Zillow Indoor Dataset (ZInD)\cite{ZInD} and Structure3D\cite{Zheng2020Structured3DAL} dataset. We also provide information regarding the training schema, used datasets, and our baselines.
\subsection{Traning}
We implement U2RLE-Net using PyTorch\cite{Paszke2019PyTorchAI}. The Adam optimizer \cite{Kingma2015AdamAM} is employed to train our network with $\beta_1 = 0.9, \beta_2 = 0.999$. The network is trained on four NVIDIA Tesla V100 GPUs for $60$ epochs on ZInD dataset and $50$ epochs on Structure3D dataset, with a batch size of 40. In addition, we set hyper-parameters in Eq. (\ref{firsttotalloss} and \ref{secondtotalloss}) as $\lambda = 10, \nu = 5$. 

Apart from the pano-stretching data augmentation described in section \ref{secondstage} for the refinement stage, we use the same data augmentation methods proposed in Horizon-Net\cite{Sun_2019_CVPR}, including left-right flipping, horizontal rotation, luminance changes, and pano-stretching during training.

\begin{figure}[t]
  \centering
  \includegraphics[width=\linewidth]{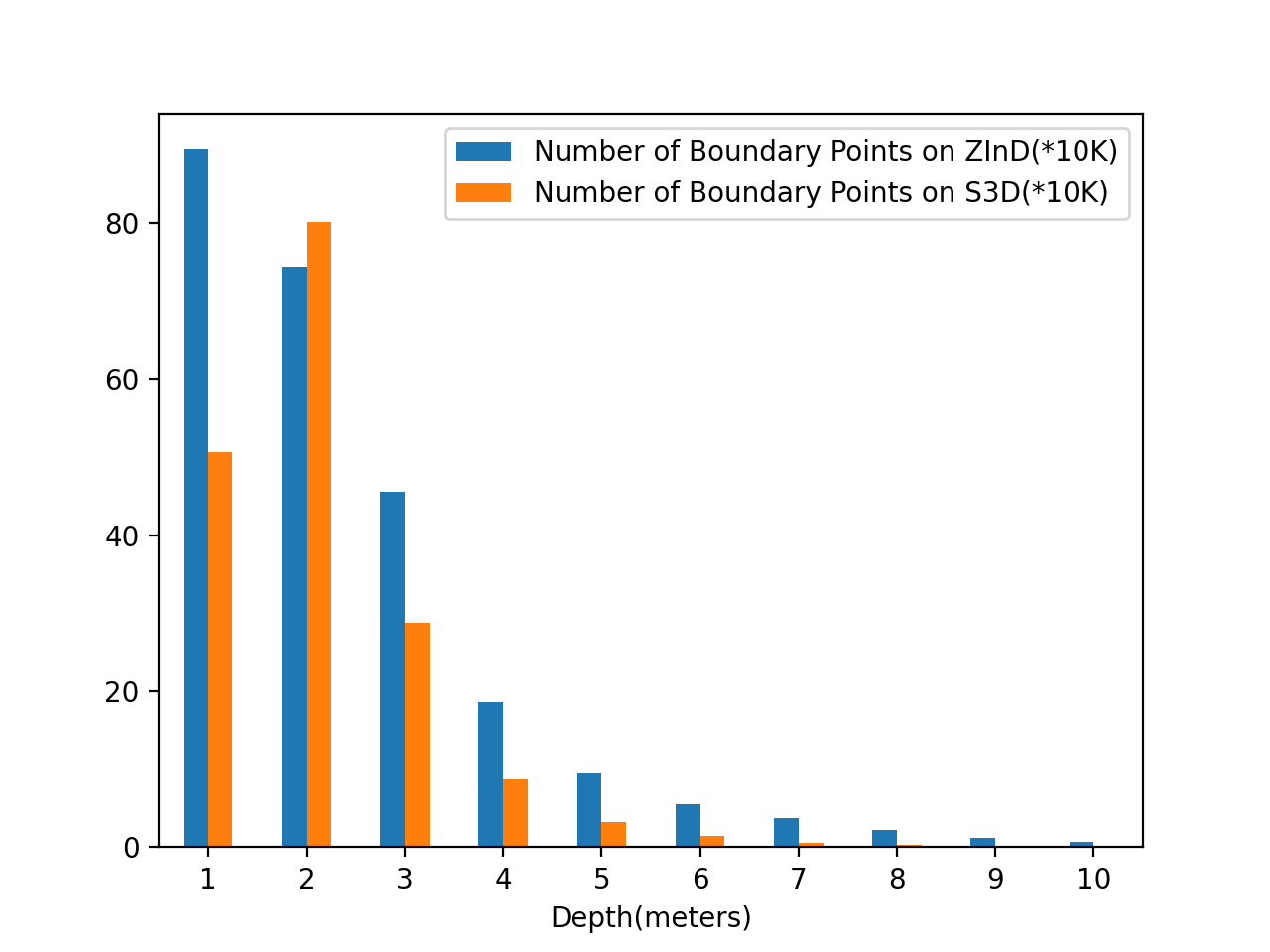}
  \caption{Depth distribution of the GT boundary points on ZInD\cite{ZInD} and Structure3D\cite{Zheng2020Structured3DAL}.}
  \label{fig:imbalance}
\end{figure}

\begin{figure*}
\begin{center}
   \includegraphics[width=0.95\linewidth]{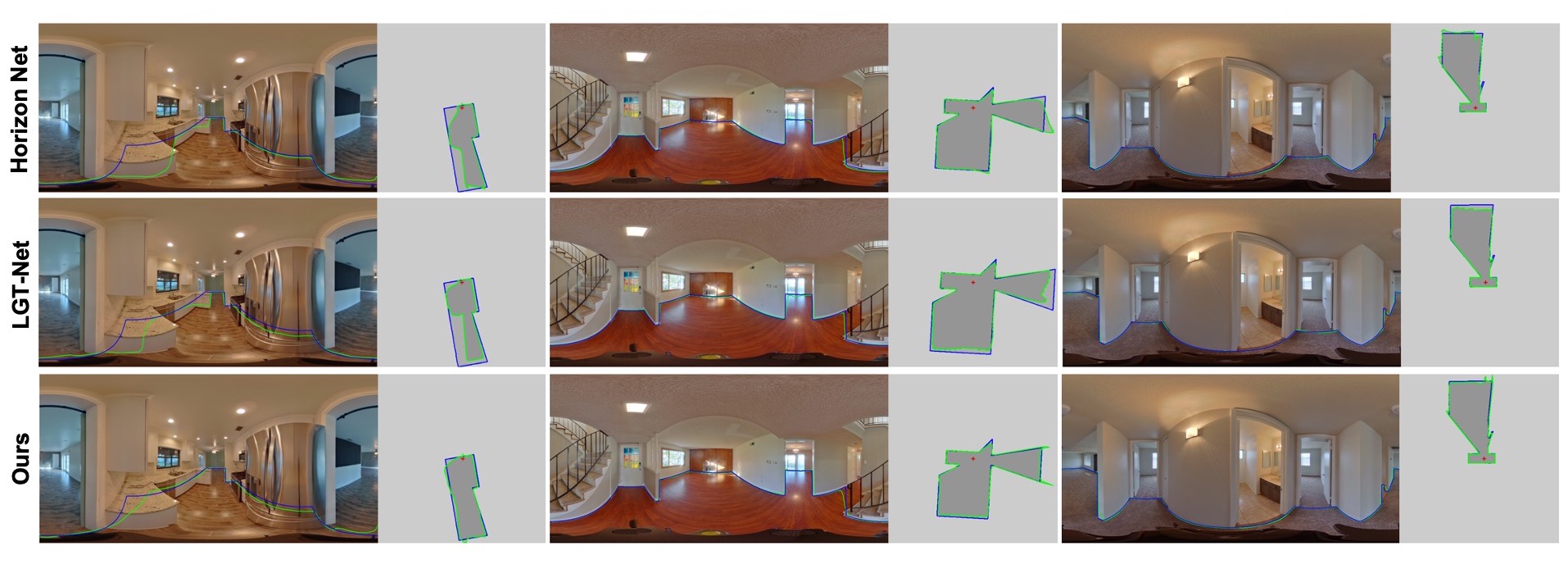}
\end{center}
\caption{Qualitative comparison on ZInD \cite{ZInD} dataset. GT layout is in \textcolor{blue}{blue} while predicted layout is in \textcolor{green}{green}. }
\label{fig:zind_qualitative}
\end{figure*}

\begin{figure*}
\begin{center}
   \includegraphics[width=0.95\linewidth]{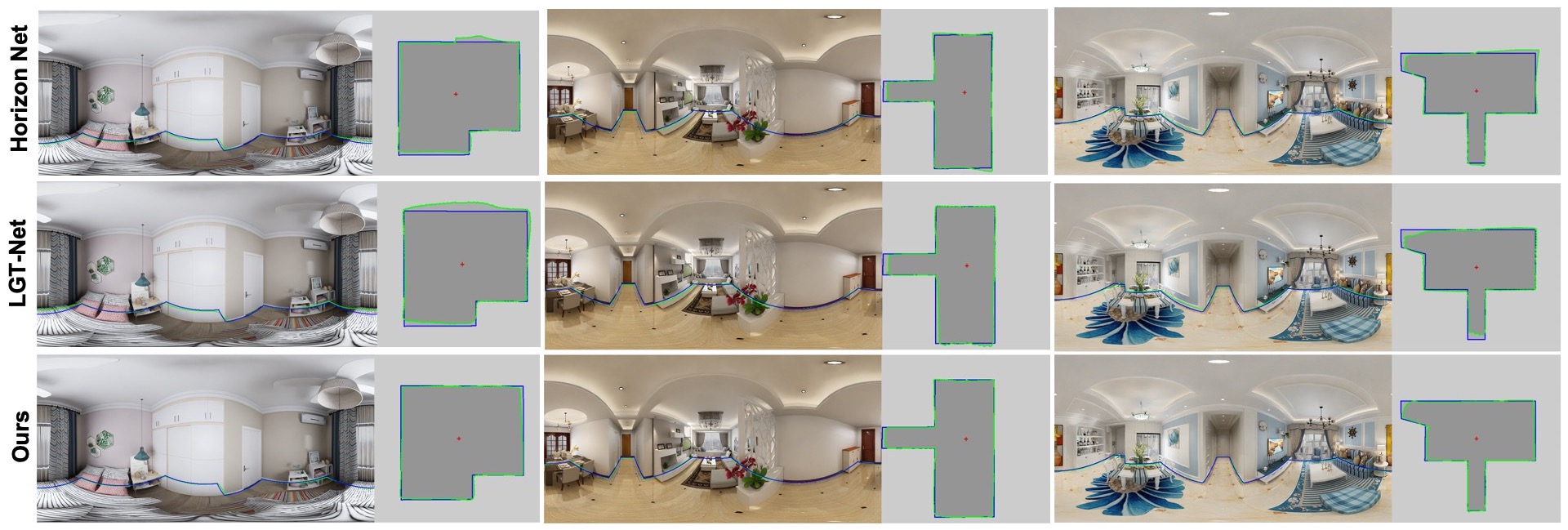}
\end{center}
\caption{Qualitative comparison on Structure3D \cite{Zheng2020Structured3DAL} dataset. GT layout is in \textcolor{blue}{blue} while predicted layout is in \textcolor{green}{green}.}
\label{fig:s3d_qualitative}
\end{figure*}

\subsection{Datasets}
Zillow indoor dataset (ZInD)\cite{ZInD} contains 71474 panoramas of 1524 unfurnished homes. It's currently the largest real world dataset with room layout annotations. It contains a close-to real world distribution of room complexities, which include a good amount of non-Manhattan walls, a wide ranges of room sizes and wall-wall corners per room. We split the dataset for training, validation, test at 20077, 2458, 2458 unique panoramas. The split is based on Horizon-Net\cite{Sun_2019_CVPR} pre-processing.\\

Structure3D\cite{Zheng2020Structured3DAL} contains more than 20K synthetic images from different rooms with 3D room layout annotations. Structure3D contains a good range of room complexity and it is furnished. The full dataset was used. The first $3000$ scenes will be used for training, $250$ scenes for validation and the remaining $250$ scenes for testing.

To better understand the imbalanced issue of the dataset, we plot the depth distribution of the boundary points on ZInD and Structure3D. The boundary point is obtained by projecting the floor-wall boundary point from image-space onto depth-space. Details about this projection can be found at section \ref{nll}. As we can see from Fig. \ref{fig:imbalance}, ZInD and Structure3D are highly imbalanced. For example, in ZInD, around 90\% of the boundary points are within 4 meters of the camera while the depth of the remaining boundary points can extend to 10+ meters.

\begin{table*}
\footnotesize
\begin{tabularx}{\textwidth}{@{} l *{14}{C} c @{}}
\toprule
\backslashbox{Model}{GT Depth (m)} 
& 1 & 2 & 3 & 4 & 5 & 6 & 7 & 8 & 9 & 10 \\ 

\midrule
Initial Stage  & \textbf{\footnotesize{0.034}} & \textbf{\footnotesize{0.045}}  &  \textbf{\footnotesize{0.067}} &  \textbf{\footnotesize{0.118}} &	 \footnotesize{0.223} &	 \footnotesize{0.330} &	\footnotesize{0.514}  &  \footnotesize{0.790} &	 \footnotesize{0.999} &	\footnotesize{1.430} \\ 


Refinement Stage  & \footnotesize{0.073} &  \footnotesize{0.063}  &  \footnotesize{0.080} &  \footnotesize{0.124} &	 \textbf{\footnotesize{0.215}} & \textbf{\footnotesize{0.278}} &	\textbf{\footnotesize{0.397}}  &	\textbf{\footnotesize{0.540}} &	\textbf{\footnotesize{0.705}} &	\textbf{\footnotesize{0.854}}\\ 


Two-stage (U2RLE)  & \footnotesize{0.034} &  \footnotesize{0.047}  &  \footnotesize{0.072} &  \footnotesize{0.127} &	 \footnotesize{0.229} & \footnotesize{0.313} &	\footnotesize{0.437}  & \footnotesize{0.614} & \footnotesize{0.776} &	\footnotesize{0.942} \\ 


\bottomrule
\end{tabularx}
\caption{Ablation study: two-stage vs single-stage.}
\label{tab:two_stage_vs_single_stage}
\end{table*} 

\begin{table*}
\footnotesize
\begin{tabularx}{\textwidth}{@{} l *{14}{C} c @{}}
\toprule
\backslashbox{Model}{GT Depth (m)} 
& 1 & 2 & 3 & 4 & 5 & 6 & 7 & 8 & 9 & 10 \\ 

\midrule
Initial Stage  & \textbf{\footnotesize{0.034}} & \textbf{\footnotesize{0.045}}  &  \textbf{\footnotesize{0.067}} &  \textbf{\footnotesize{0.118}} &	 \textbf{\footnotesize{0.223}} &	 \textbf{\footnotesize{0.330}} &	\textbf{\footnotesize{0.514} } &  \textbf{\footnotesize{0.790}} &	 \textbf{\footnotesize{0.999}} &	\textbf{\footnotesize{1.430} }\\ 


Initial Stage Without CPHC module & \footnotesize{0.042} & \footnotesize{0.055}  & \footnotesize{0.080} & \footnotesize{0.153} &	\footnotesize{0.291} &	\footnotesize{0.474} &	\footnotesize{0.724}  &	\footnotesize{1.121} &	\footnotesize{1.151} &	\footnotesize{2.086}  \\ 


\hline

Refinement Stage  & \footnotesize{0.073} &  \textbf{\footnotesize{0.063}}  &  \footnotesize{0.080} &  \textbf{\footnotesize{0.124}} &	 \textbf{\footnotesize{0.215} }& \textbf{\footnotesize{0.278}} &	\textbf{\footnotesize{0.397} } &	\textbf{\footnotesize{0.540}} &	\textbf{\footnotesize{0.705}} &	\textbf{\footnotesize{0.854}}\\ 


Refinement Stage Without CPHC module  & \textbf{\footnotesize{0.067}} & \footnotesize{0.066}  &  \textbf{\footnotesize{0.078}} &  \footnotesize{0.128} &	 \footnotesize{0.227} &	 \footnotesize{0.315} &	\footnotesize{0.458}  &  \footnotesize{0.622} &	 \footnotesize{0.785} &	\footnotesize{1.056}  \\ 


\bottomrule
\end{tabularx}
\caption{Ablation study: proposed CPHC module.}
\label{tab:new_hcm}
\end{table*}

\subsection{Baselines}
Currently, none of the models have been trained on the "visible geometry" of ZInD. Therefore, we retrained all the models based on their repository on the ZinD dataset. Additionally, if needed, baselines are also trained on Structure3D. We compare our U2RLE with baselines built upon recent state-of-the-art (SOTA) layout estimation methods: HorizonNet\cite{Sun_2019_CVPR}, LED2-Net\cite{Wang_2021_CVPR} and, LGT-Net\cite{jiang2022lgt}. Also, due to the complexity of the structures in "visible geometry", post processing techniques introduced in \cite{Yang:2019:DuLa-Net,Sun_2019_CVPR} do not perform effectively and can significantly drop the performance. Therefore, for evaluation purposes, we evaluate all the models without the post-processing techniques.  

\subsection{Evaluation Metrics}
In order to show the effectiveness of our model, we reported our error using two metrics, mean depth error and 2D IoU. 2D IoU shows the overall performance of the models, but it does not explicitly show the effectiveness of predictions based on distance. As shown in Figure \ref{fig:motivation}, the error on the distant regions will be amplified and can degrade 2D IoU significantly. To better study the prediction errors, we evaluate all of the models based on the mean depth error. To calculate the error, first the predicted and ground truth boundary are projected to top-down view. The distance from the projected ground truth boundary to the origin is used to calculate the error. The difference between the predicted and ground truth boundary is defined as the depth error. To show the overall model's performance, we also report 2D IoU. 


\begin{table*}
\footnotesize
\begin{tabularx}{\textwidth}{@{} l *{14}{C} c @{}}
\toprule
\backslashbox{Model}{GT Depth (m)} 
& 1 & 2 & 3 & 4 & 5 & 6 & 7 & 8 & 9 & 10 \\ 

\midrule
Initial Stage  & \textbf{\footnotesize{0.034}} & \textbf{\footnotesize{0.045}}  &  \textbf{\footnotesize{0.067}} &  \textbf{\footnotesize{0.118}} &	 \textbf{\footnotesize{0.223}} &	 \textbf{\footnotesize{0.330}} &	\footnotesize{0.514}  &  \footnotesize{0.790} &	 \footnotesize{0.999} &	\footnotesize{1.430} \\ 


Without Uncertainty & \footnotesize{0.035} & \footnotesize{0.046}  & \footnotesize{0.069} & \footnotesize{0.123} &	\footnotesize{0.233} &	\footnotesize{0.332} &	\textbf{\footnotesize{0.495} } &	\textbf{\footnotesize{0.703}} &	\textbf{\footnotesize{0.926}} &	\textbf{\footnotesize{1.242}}  \\ 


\bottomrule
\end{tabularx}
\caption{Ablation study: uncertainty prediction.}
\label{tab:uncertainty_prediction}
\end{table*} 

\begin{table*}
\footnotesize
\begin{tabularx}{\textwidth}{@{} l *{14}{C} c @{}}
\toprule
\backslashbox{Model}{GT Depth (m)} 
& 1 & 2 & 3 & 4 & 5 & 6 & 7 & 8 & 9 & 10 \\ 

\midrule

Refinement Stage  & \footnotesize{0.073} &  \footnotesize{0.063}  &  \textbf{\footnotesize{0.080}} &  \textbf{\footnotesize{0.124}} &	 \textbf{\footnotesize{0.215}} & \textbf{\footnotesize{0.278}} &	\textbf{\footnotesize{0.397}}  &	\textbf{\footnotesize{0.540}} &	\textbf{\footnotesize{0.705}} &	\textbf{\footnotesize{0.854}}\\ 


Without Distance-aware Loss  & \textbf{\footnotesize{0.073}} &  \textbf{\footnotesize{0.063} } &  \footnotesize{0.081} &  \footnotesize{0.135} &	 \footnotesize{0.240} &	 \footnotesize{0.354} & \footnotesize{0.546}  &  \footnotesize{0.746} &	 \footnotesize{0.929} &	 \footnotesize{1.444}   \\ 


\bottomrule
\end{tabularx}
\caption{Ablation study: distance-aware loss.}
\label{tab:distance_aware_loss}
\end{table*}

\subsection{Quantitative Evaluation}
Our evaluation on ZInD\cite{ZInD}, is shown in Table \ref{tab:zind}. All the errors are reported in meters. Our approach has lower mean depth error on distant regions in comparison to all other approaches. Also, the model can outperform the available models in the overall 2D IoU. To show the effectiveness of our model in furnished settings, we compare our approach to other baselines. Table \ref{tab:s3d} shows the introduced metrics on Structure3D\cite{Zheng2020Structured3DAL} dataset. Our approach can outperform significantly on almost all close and distant regions. 

\subsection{Qualitative Evaluation}
Figure \ref{fig:zind_qualitative} shows some examples on ZInD dataset\cite{ZInD} and the predicted boundary both on panorama and top-down view. The \textcolor{green}{green} color represents the predicted floor boundary and \textcolor{blue}{blue} indicates the Ground Truth. The first image is an example where our model is capable of capturing both distant and close regions appropriately. The second image illustrates the effectiveness of our model in large open spaces. Finally, the last image represents a complex structure that our model is able to predict more accurately.

Figure \ref{fig:s3d_qualitative} shows the results on Structure3D\cite{Zheng2020Structured3DAL} dataset. Compared to ZinD\cite{ZInD}, Structure3D is furnished. First image shows the results of our model compared to other baselines when occlusions occur in close regions, where our model is able to handle the occlusion more effectively. Additionally, we provided examples of large open areas in the second and third images.

\section{Ablation Study}
\label{ablation}
We conduct experiments to determine how individual components affect the U2RLE architecture. All the ablation studies are conducted on ZInD dataset. In particular, we examine the following variants:

\begin{itemize}
    \item Compare the proposed two-stage model with single-stage model.
    \item Replace the proposed Channel-Preserving Height Compression (CPHC) Module (Section \ref{hcm}) with the original height compression module proposed by \cite{Sun_2019_CVPR}. 
    \item Replace the uncertainty prediction in the initial stage with $\ell_1$ loss function to predict the boundaries directly.
    \item Finally, we do experiments on the effect of our distance-aware loss (Section \ref{distant_aware_loss}) in the refinement stage.
    
\end{itemize}

\subsection{Two-stage VS Single-stage}
The motivation of designing our two-stage model is that current single stage models cannot achieve optimal performance for both close walls and distant walls.  As we can see from Tab. \ref{tab:two_stage_vs_single_stage}, the initial stage model works well for close walls where depth ranges from 1m to 4m while while the refinement stage performs better on distant walls where depth ranges from 5m to 10m. With our proposed two-stage architecture (U2RLE), we can merge the results from the initial stage and the refinement stage to achieve good performance for both close walls and distant walls. 

One may argue that we can merge based on a depth threshold like 4.5m. In reality, we do not know the GT depth. The predicted depth is pretty noisy. Directly merging based on the predicted depth will result in sub-optimal result. In our merging module, we use the predicted uncertainty score to guide the merging and utilize the predicted depth to reduce outlier, which gives better result.

\subsection{Proposed CPHC Module}
In our U2RLE, a CPHC module is proposed for the initial stage and the refinement stage to compress information along height dimension. As we can see from Tab. \ref{tab:new_hcm}, the mean depth error of the initial stage becomes worse for all depths if we replace the proposed CPHC module with the original height compression module. Similarly, the proposed CPHC module helps to achieve better depth estimation for the refinement stage, especially for the distant walls. These results verify the effectiveness of the proposed height compression module.

\subsection{Uncertainty Prediction}
The initial stage predicts both the floor-wall boundary and uncertainty score. The uncertainty prediction is not only important for the merging step but also can improve the performance. As we can see from Tab. \ref{tab:uncertainty_prediction}, the uncertainty prediction helps to improve the performance for close walls. One may also notice that our initial stage with uncertainty prediction has worse results for distant walls. Since the initial stage is designed to predict close walls, this will not affect the final result. This also supports our motivation of the two-stage model where we can use different loss functions for the initial stage and the refinement stage to handle close walls and distant walls.

The uncertainty prediction is optimized via negative log likelihood loss, which can be minimized in two ways for far away regions, by predicting an accurate boundary ($ \mu $) or by increasing the uncertainty ($ \sigma $) for those far away regions. Our model has decided to minimize the loss by increasing the uncertainty there, which means that the optimizer do not need to find an accurate $ \mu $ for far away regions. In this way, the initial stage can allocate more learning capacity on close walls.

\subsection{Distance-aware Loss}
In the refinement stage, a distance-aware loss is proposed to force the model focus more on the distant walls. As we can see from Tab. \ref{tab:distance_aware_loss}, the proposed distance-aware loss significantly reduces the mean depth error for distant walls.

\section{Conclusions and Future Work}
We presented effective improvement of the state-of-the-art in 1D layout estimation from a single panoramic view. 
With similar motivation to traditional depth estimation methods from perspective views, which suffers from lower quality estimates for far away regions, our approach focused on improvements for distant boundaries. We proposed a novel CPHC model to improve the overall accuracy with floor-wall boundary confidence, introduced a refined stage to produce higher accuracy for distance regions and apply the uncertainty guided merging module for final predictions. We presented comprehensive evaluation along the ablation study showing the effect of different improved components. 


Current generation of high-resolution panoramas can be easily 8K at full resolution. It is computationally expensive to use the full resolution for a single stage network and the reduction of resolution affects more
the quality of the estimates in far away regions. We believe that the two-stage approach can provide benefit in this setting;  lower resolution data can be used for initial stage, and the higher resolution is used only for areas with high uncertainty. Testing the effectiveness of the proposed model in the multi-resolution setting can be explored in the future work. 



{\small
\bibliographystyle{ieee_fullname}
\bibliography{latex/final}

\begin{thebibliography}{10}\itemsep=-1pt

\bibitem{armeni2017joint}
Iro Armeni, Sasha Sax, Amir~R Zamir, and Silvio Savarese.
\newblock Joint 2d-3d-semantic data for indoor scene understanding.
\newblock {\em arXiv preprint arXiv:1702.01105}, 2017.

\bibitem{Bae_IronDepth_arxiv2022}
Gwangbin Bae, Ignas Budvytis, and Roberto Cipolla.
\newblock Irondepth: Iterative refinement of single-view depth using surface
  normal and its uncertainty.
\newblock In {\em DOI: 10.48550/arXiv.2210.03676}, Obtober 2022.

\bibitem{Chen2021WideBaselineRC}
Kefan Chen, Noah Snavely, and Ameesh Makadia.
\newblock Wide-baseline relative camera pose estimation with directional
  learning.
\newblock {\em 2021 IEEE/CVF Conference on Computer Vision and Pattern
  Recognition (CVPR)}, pages 3257--3267, 2021.

\bibitem{ZInD}
Steve Cruz, Will Hutchcroft, Yuguang Li, Naji Khosravan, Ivaylo Boyadzhiev, and
  Sing~Bing Kang.
\newblock Zillow indoor dataset: Annotated floor plans with 360º panoramas and
  3d room layouts.
\newblock In {\em Proceedings of the IEEE/CVF Conference on Computer Vision and
  Pattern Recognition (CVPR)}, pages 2133--2143, June 2021.

\bibitem{flint2010dynamic}
Alex Flint, Christopher Mei, David Murray, and Ian Reid.
\newblock A dynamic programming approach to reconstructing building interiors.
\newblock In {\em European conference on computer vision}, pages 394--407.
  Springer, 2010.

\bibitem{He2016DeepRL}
Kaiming He, X. Zhang, Shaoqing Ren, and Jian Sun.
\newblock Deep residual learning for image recognition.
\newblock {\em 2016 IEEE Conference on Computer Vision and Pattern Recognition
  (CVPR)}, pages 770--778, 2016.

\bibitem{hedau2009recovering}
Varsha Hedau, Derek Hoiem, and David Forsyth.
\newblock Recovering the spatial layout of cluttered rooms.
\newblock In {\em 2009 IEEE 12th international conference on computer vision},
  pages 1849--1856. IEEE, 2009.

\bibitem{jiang2022lgt}
Zhigang Jiang, Zhongzheng Xiang, Jinhua Xu, and Ming Zhao.
\newblock Lgt-net: Indoor panoramic room layout estimation with geometry-aware
  transformer network.
\newblock In {\em Proceedings of the IEEE Conference on Computer Vision and
  Pattern Recognition (CVPR)}, 2022.

\bibitem{Kingma2015AdamAM}
Diederik~P. Kingma and Jimmy Ba.
\newblock Adam: A method for stochastic optimization.
\newblock {\em CoRR}, abs/1412.6980, 2015.

\bibitem{lee2017roomnet}
Chen-Yu Lee, Vijay Badrinarayanan, Tomasz Malisiewicz, and Andrew Rabinovich.
\newblock Roomnet: End-to-end room layout estimation.
\newblock In {\em Proceedings of the IEEE international conference on computer
  vision}, pages 4865--4874, 2017.

\bibitem{Loquercio_Uncertainty_RAL2020}
Antonio Loquercio, Mattia Segù, and Davide Scaramuzza.
\newblock A general framework for uncertainty estimation in deep learning.
\newblock In {\em IEEE Robotics and Automation Letters, Vol. 5, No. 2}, April
  2020.

\bibitem{Mathew2020ReviewOR}
Bincy~P Mathew.
\newblock Review on room layout estimation from a single image.
\newblock {\em International Journal of Engineering Research and}, 9, 2020.

\bibitem{Paszke2019PyTorchAI}
Adam Paszke, Sam Gross, Francisco Massa, Adam Lerer, James Bradbury, Gregory
  Chanan, Trevor Killeen, Zeming Lin, Natalia Gimelshein, Luca Antiga, Alban
  Desmaison, Andreas K{\"o}pf, Edward Yang, Zach DeVito, Martin Raison, Alykhan
  Tejani, Sasank Chilamkurthy, Benoit Steiner, Lu Fang, Junjie Bai, and Soumith
  Chintala.
\newblock Pytorch: An imperative style, high-performance deep learning library.
\newblock {\em ArXiv}, abs/1912.01703, 2019.

\bibitem{Pintore2020AtlantaNetIT}
Giovanni Pintore, Marco Agus, and E. Gobbetti.
\newblock Atlantanet: Inferring the 3{D} indoor layout from a single 360-deg
  image beyond the {M}anhattan world assumption.
\newblock In {\em ECCV}, 2020.

\bibitem{ren2021bmse}
Jiawei Ren, Mingyuan Zhang, Cunjun Yu, and Ziwei Liu.
\newblock Balanced mse for imbalanced visual regression.
\newblock In {\em Proceedings of the IEEE/CVF Conference on Computer Vision and
  Pattern Recognition}, 2022.

\bibitem{ren2016coarse}
Yuzhuo Ren, Shangwen Li, Chen Chen, and C-C~Jay Kuo.
\newblock A coarse-to-fine indoor layout estimation (cfile) method.
\newblock In {\em Asian conference on computer vision}, pages 36--51. Springer,
  2016.

\bibitem{solarte2022mlc}
Bolivar Solarte, Chin-Hsuan Wu, Yueh-Cheng Liu, Yi-Hsuan Tsai, and Min Sun.
\newblock 360-mlc: Multi-view layout consistency for self-training and
  hyper-parameter tuning.
\newblock In {\em Advances in Neural Information Processing Systems}, 2022.

\bibitem{gprnet_2022}
Jheng{-}Wei Su, Chi{-}Han Peng, Peter Wonka, and Hung{-}Kuo Chu.
\newblock Gpr-net: Multi-view layout estimation via a geometry-aware panorama
  registration network.
\newblock {\em CoRR}, abs/2210.11419, 2022.

\bibitem{Sun_2019_CVPR}
Cheng Sun, Chi-Wei Hsiao, Min Sun, and Hwann-Tzong Chen.
\newblock Horizonnet: Learning room layout with 1d representation and pano
  stretch data augmentation.
\newblock In {\em The IEEE Conference on Computer Vision and Pattern
  Recognition (CVPR)}, June 2019.

\bibitem{SunSC21}
Cheng Sun, Min Sun, and Hwann{-}Tzong Chen.
\newblock Hohonet: 360 indoor holistic understanding with latent horizontal
  features.
\newblock In {\em CVPR}, 2021.

\bibitem{Wang_2021_CVPR}
Fu-En Wang, Yu-Hsuan Yeh, Min Sun, Wei-Chen Chiu, and Yi-Hsuan Tsai.
\newblock Led2-net: Monocular 360deg layout estimation via differentiable depth
  rendering.
\newblock In {\em Proceedings of the IEEE/CVF Conference on Computer Vision and
  Pattern Recognition (CVPR)}, pages 12956--12965, June 2021.

\bibitem{wang2022psmnet}
Haiyan Wang, Will Hutchcroft, Yuguang Li, Zhiqiang Wan, Ivaylo Boyadzhiev,
  Yingli Tian, and Sing~Bing Kang.
\newblock Psmnet: Position-aware stereo merging network for room layout
  estimation.
\newblock In {\em Proceedings of the IEEE/CVF Conference on Computer Vision and
  Pattern Recognition}, pages 8616--8625, 2022.

\bibitem{yan20203d}
Chenggang Yan, Biyao Shao, Hao Zhao, Ruixin Ning, Yongdong Zhang, and Feng Xu.
\newblock 3d room layout estimation from a single rgb image.
\newblock {\em IEEE Transactions on Multimedia}, 22(11):3014--3024, 2020.

\bibitem{Yang:2019:DuLa-Net}
Shang-Ta Yang, Fu-En Wang, Chi-Han Peng, Peter Wonka, Min Sun, and Hung-Kuo
  Chu.
\newblock Dula-net: {A} dual-projection network for estimating room layouts
  from a single {RGB} panorama.
\newblock In {\em {IEEE} Conference on Computer Vision and Pattern Recognition,
  {CVPR} 2019}, pages 3363--3372, 2019.

\bibitem{zhang2014panocontext}
Yinda Zhang, Shuran Song, Ping Tan, and Jianxiong Xiao.
\newblock Panocontext: A whole-room 3d context model for panoramic scene
  understanding.
\newblock In {\em European conference on computer vision}, pages 668--686.
  Springer, 2014.

\bibitem{zhao20223d}
Yining Zhao, Chao Wen, Zhou Xue, and Yue Gao.
\newblock 3d room layout estimation from a cubemap of panorama image via deep
  manhattan hough transform.
\newblock In {\em European Conference on Computer Vision}, pages 637--654.
  Springer, 2022.

\bibitem{Zheng2020Structured3DAL}
Jia Zheng, Junfei Zhang, Jing Li, Rui Tang, Shenghua Gao, and Zihan Zhou.
\newblock Structured3d: A large photo-realistic dataset for structured 3d
  modeling.
\newblock In {\em ECCV}, 2020.

\bibitem{zou2018layoutnet}
Chuhang Zou, Alex Colburn, Qi Shan, and Derek Hoiem.
\newblock Layoutnet: Reconstructing the 3d room layout from a single rgb image.
\newblock In {\em Proceedings of the IEEE Conference on Computer Vision and
  Pattern Recognition}, pages 2051--2059, 2018.

\end{thebibliography}
}

\pagebreak
\clearpage
\begin{center}
\textbf{U2RLE: Uncertainty-Guided 2-Stage Room Layout Estimation\\(Supplementary)}
\end{center}

\setcounter{equation}{0}
\setcounter{figure}{0}
\setcounter{table}{0}
\setcounter{page}{1}
\makeatletter
\renewcommand{\theequation}{S\arabic{equation}}
\renewcommand{\thefigure}{S\arabic{figure}}

\section{Imbalanced Learning}


Since the dataset is highly imbalanced, one potential solution is to combine the SOTA layout estimation model with the SOTA data re-balancing method. We use the SOTA balanced MSE proposed in \cite{ren2021bmse} to alleviate the issue in our task. We compared our two-stage model with two baselines: SOTA LGT-Net + Balanced MSE \cite{ren2021bmse} and our initial stage + Balanced MSE \cite{ren2021bmse}. The results in Table~\ref{tab:imbalanced} show that a single-stage model with SOTA re-balancing method does not work well on this challenging problem.

\begin{table}[ht]
\centering
\resizebox{0.4\textwidth}{!}{

\begin{tabular}[t]{lcc}
\hline
 \small{Model}& \small{2$\mathrm{DIoU}$}\\
\hline
 \footnotesize{LGT-Net\cite{jiang2022lgt} + Balanced MSE \cite{ren2021bmse}} & \footnotesize{82.21\%} \\
 \footnotesize{Our Initial Stage + Balanced MSE \cite{ren2021bmse}}& \footnotesize{70.38\%} \\
 \footnotesize{Ours(U2RLE)} & \footnotesize{91.39\%} \\

\hline
\end{tabular}
}
\caption{Quantitative results comparing our two-stage model with single-stage model + SOTA re-balancing technique.}
\label{tab:imbalanced}
\end{table}

%


\begin{figure*}[t]
  \centering
  \includegraphics[width=\linewidth]{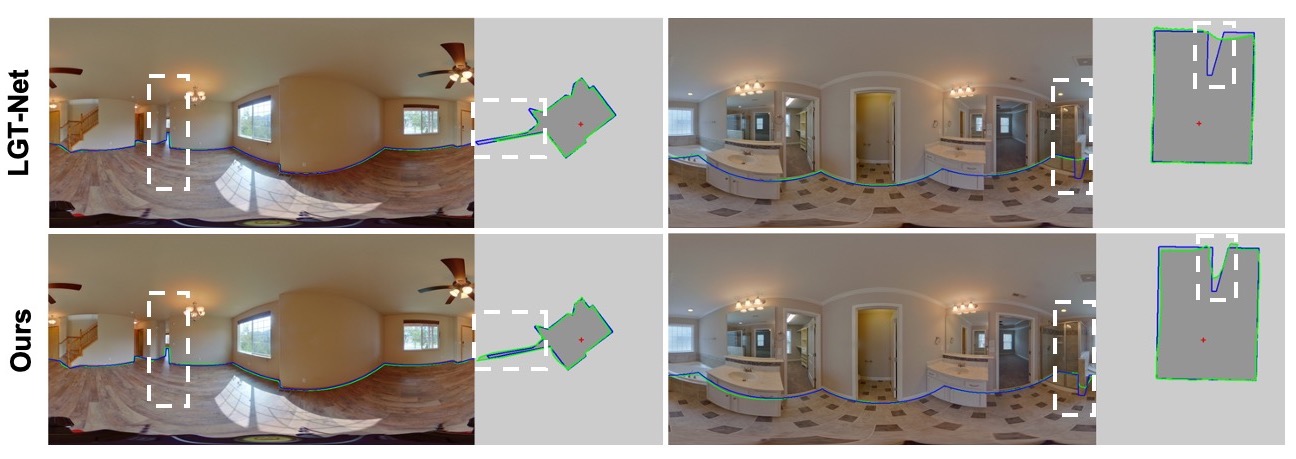}
  \caption{Examples of when the models fail to predict sharp changes.}
  \label{fig:imbalance}
\end{figure*}

\begin{figure*}[t]
  \centering
  \includegraphics[width=\linewidth]{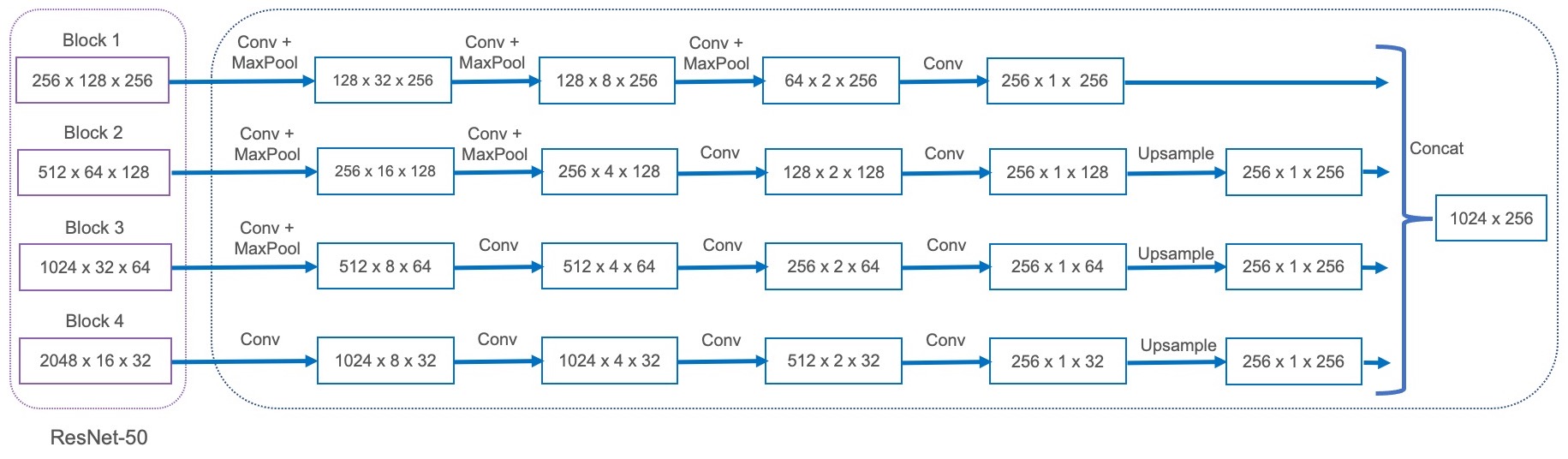}
  \caption{The architecture of the proposed channel-preserving height compression (CPHC) module.}
  \label{fig:cphc}
\end{figure*}

\section{Limitations}

\subsection{Post-Processing}




Our proposed model did not use the post-processing proposed in HorizonNet\cite{Sun_2019_CVPR}. HorizonNet's post-processing assumes that the room only contains Manhattan walls. Non-Manhattan walls are pretty common, especially in ZInD's\cite{ZInD} ``visible-geometry". A new post-processing that can handle non-Manhattan walls is needed. We leave this for our future work.

\subsection{Sharp Depth Discontinuity}


Our proposed model does not work well on predicting sharp depth discontinuity over a small area. Some exapmles are shown in Figure \ref{fig:imbalance} This is because the features from ResNet's block 1 $ \sim $ 4 have a large receptive field. Based on these features, we can only get smooth predictions.

\section{Channel-preserving Height Compression Module}
The architecture of the proposed channel-preserving height compression (CPHC) module is shown in Figure~\ref{fig:cphc}. The input to CPHC module is the features from ResNet-50\cite{He2016DeepRL}. Features from blocks 1-4 are passed through different branches in CPHC module. After convolution, pooling, and up-sampling, the height dimension is compressed, and the dimension of these features becomes $ 256 \times 1 \times 256 $. Finally, these features are concatenated along channel dimension.

\section{More Results}
\begin{figure*}[t]
  \centering
  \includegraphics[width=\linewidth]{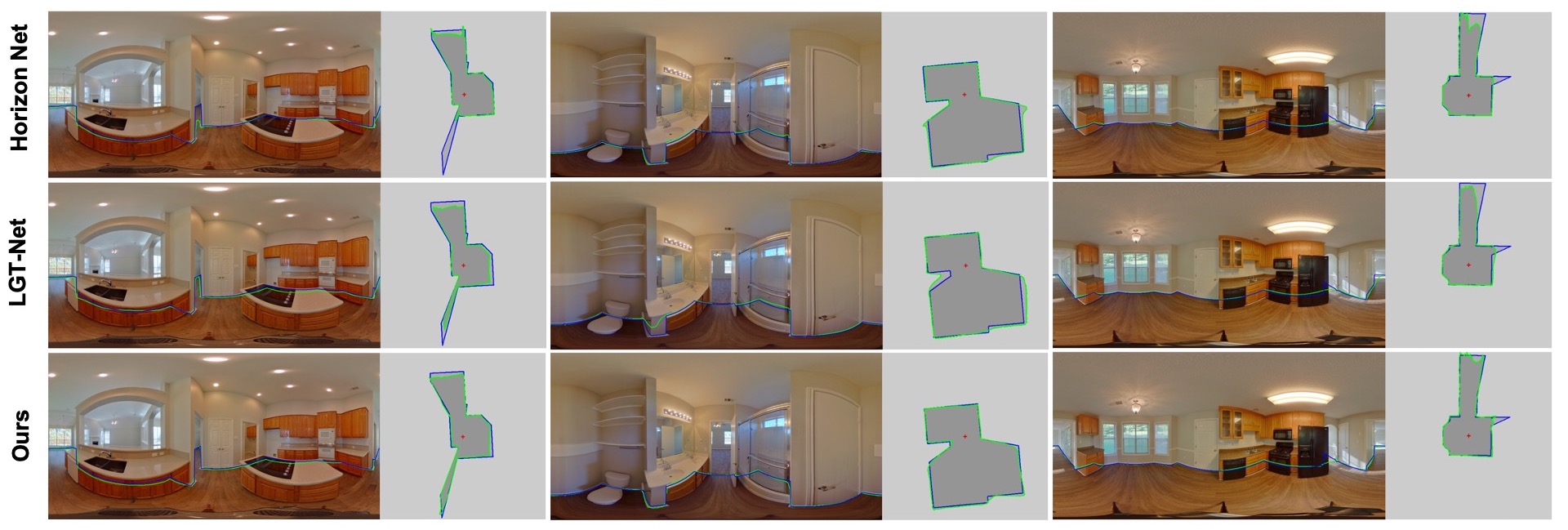}
  \includegraphics[width=\linewidth]{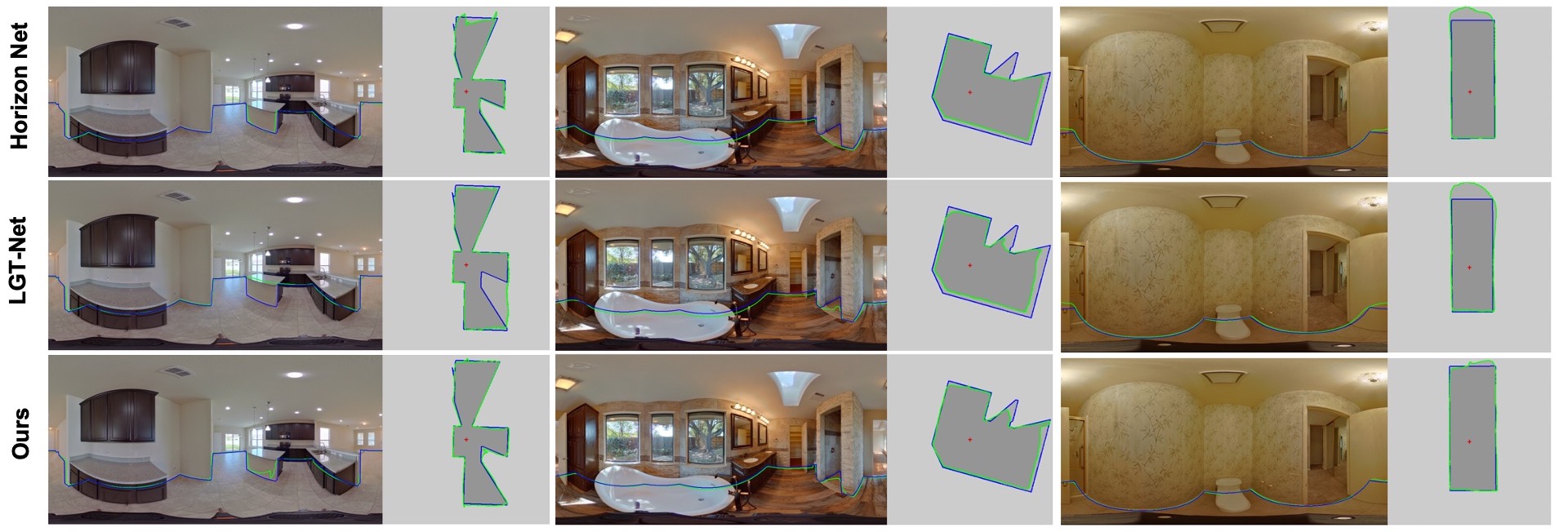}
  \includegraphics[width=\linewidth]{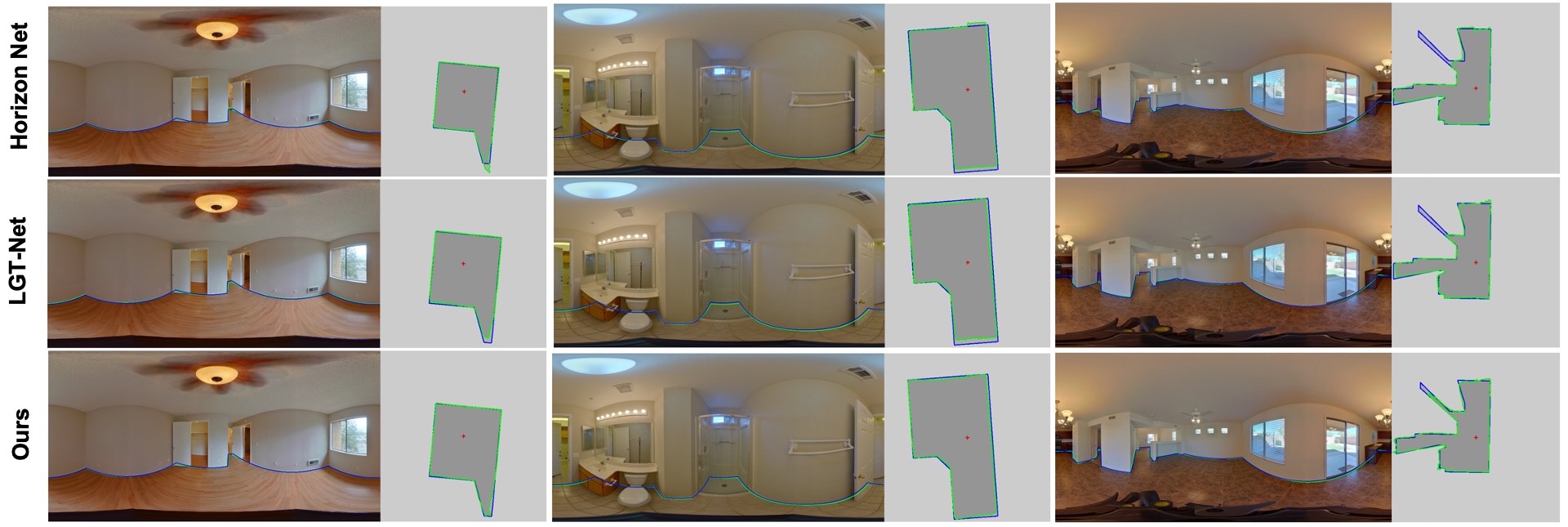}

   \caption{More qualitative comparison on ZInD \cite{ZInD} dataset. GT layout is in \textcolor{blue}{blue} while predicted layout is in \textcolor{green}{green}.}
  \label{fig:morezind}
\end{figure*}

\begin{figure*}[t]
  \centering
  \includegraphics[width=\linewidth]{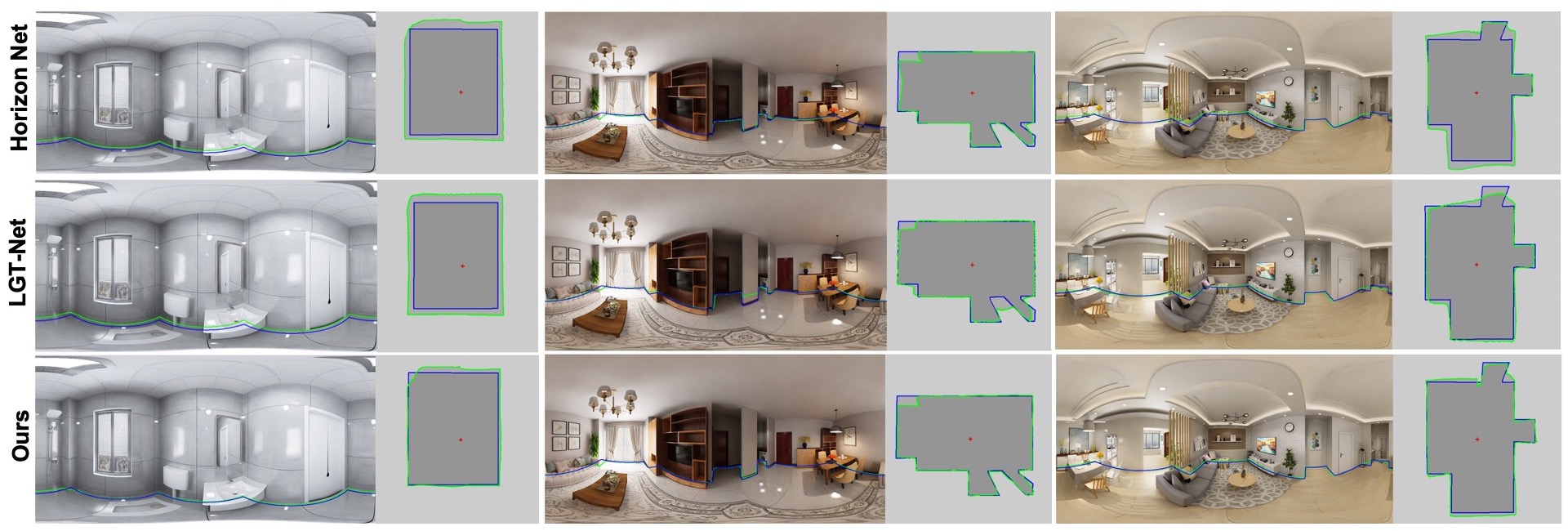}
  \includegraphics[width=\linewidth]{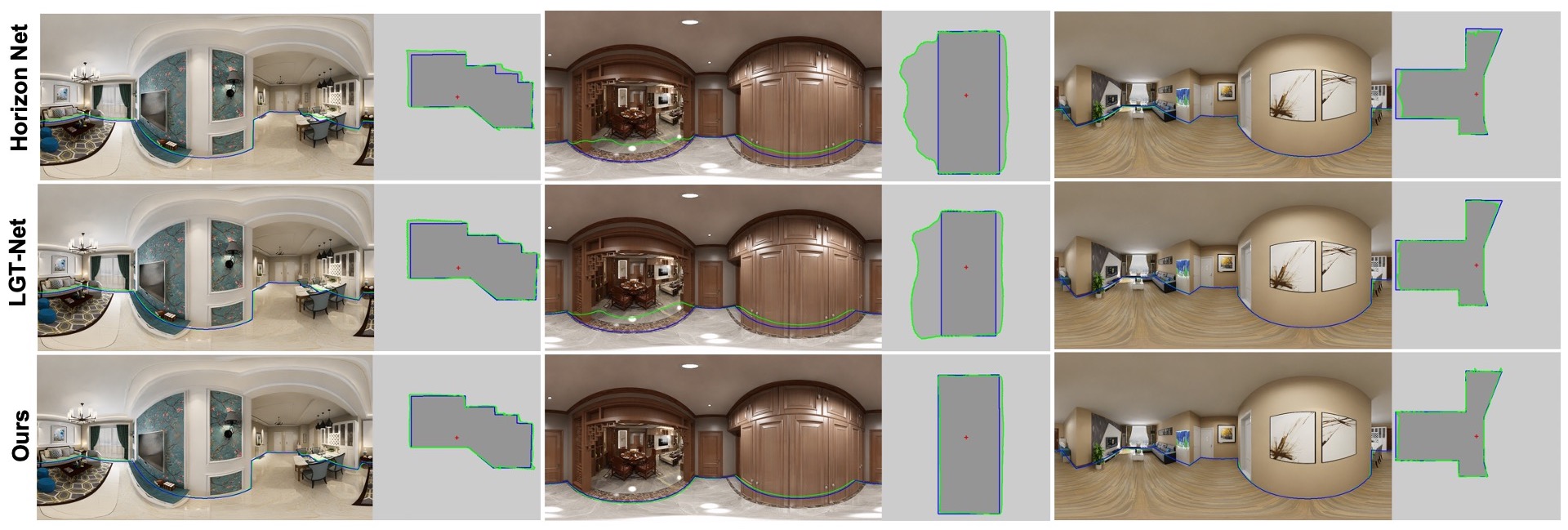}
  \includegraphics[width=\linewidth]{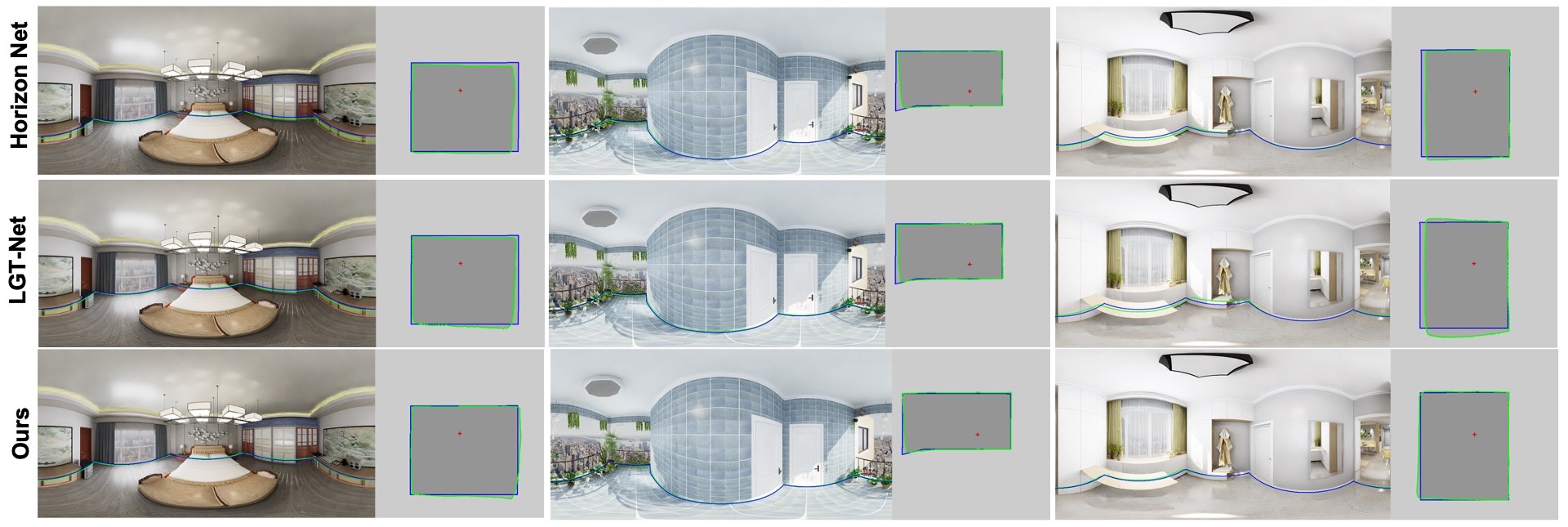}
   \caption{More qualitative comparison on Structure3D \cite{Zheng2020Structured3DAL} dataset. GT layout is in \textcolor{blue}{blue} while predicted layout is in \textcolor{green}{green}.}
  \label{fig:mores3d}
\end{figure*}

In order to better compare our proposed method with other models, we provide more qualitative results on ZInD\cite{ZInD} and Structure3D\cite{Zheng2020Structured3DAL} datasets in Figures~\ref{fig:morezind} and \ref{fig:mores3d}, respectively. 


\section{Failure Cases}
\begin{figure*}[t]
  \centering
  \includegraphics[width=\linewidth]{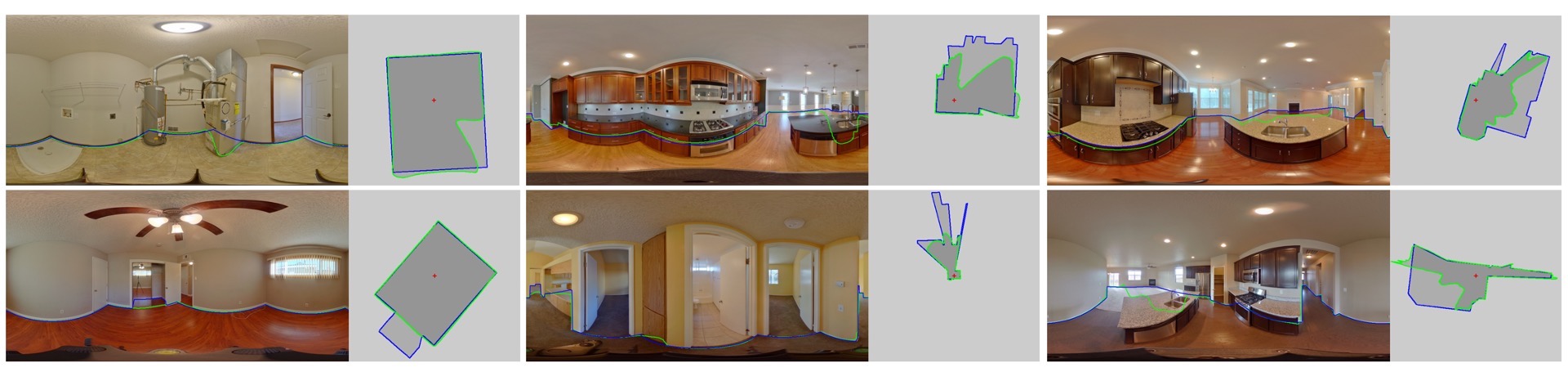}
   \caption{Further failure cases of our approach. The first column shows some cases where the floor boundary is highly occluded. In the second and third columns, the model confuses the kitchen island with the actual floor boundary and fails to predict the actual floor boundary.}
  \label{fig:fail}
\end{figure*}

In this section, we provide some other examples of when our system fails. Typical failures occur when a kitchen island is present or the floor is heavily obstructed. Figure~\ref{fig:fail} shows some representative examples.

\end{document}